\documentclass[sigconf,nonacm]{acmart}
\usepackage{graphicx}
\usepackage{tabularx}
\usepackage{booktabs}

\setcopyright{none}
\acmConference[SAO @ CAIS'26]{SAO Workshop at CAIS'26}{May 26, 2026}{San Jose, California, USA}
\acmBooktitle{SAO Workshop at CAIS'26, May 26, 2026, San Jose, California, USA}
\AtBeginDocument{%
  \pagestyle{fancy}%
  \fancyhf{}% Clear default headers and footers
  \fancyhead[L]{SAO Workshop at CAIS'26, May 26, 2026, San Jose, California, USA}%
  \fancyhead[R]{J. Axisa}%
  \fancyfoot[C]{\thepage}%
  \providecommand\BibTeX{{%
    \normalfont B\kern-.05em{\sc i\kern-.025em b}\kern-.08em T\kern-.1667em\lower.7ex\hbox{E}\kern-.125emX}}}

\begin{document}

\title{Architecting Conversational Data Systems for Stateless LLM APIs}
\subtitle{The Hydration Proxy Pattern}

\author{Joseph Axisa}
\email{josephaxisa@google.com}
\affiliation{%
  \institution{Google Cloud}
  \city{California}
  \country{USA}
}

\renewcommand{\shortauthors}{J. Axisa}

\begin{abstract}
As enterprise platforms transition to conversational reasoning interfaces, the stateless nature of high-performance LLM APIs creates a fundamental architectural gap. While statelessness enables massive horizontal scalability for AI providers, it forces the consuming application to manage the entire burden of conversational state and semantic memory. This work identifies the \textit{Hydration Proxy Pattern}, a three-tier architecture synthesized from the operational requirements of production-grade analytical platforms. By decoupling session persistence from the reasoning engine, the framework ensures platform sovereignty over conversational data while enabling secure, multi-stage semantic grounding. We further propose the \textit{Context Stabilization Mandate} to resolve the tension between sovereign state management and infrastructure-level KV caching.
\end{abstract}

\maketitle
\thispagestyle{fancy}

\section{The Challenge: The Stateless Paradox}
The integration of LLMs into professional data environments is complicated by the \textit{Stateless Paradox}. Analytical exploration is an inherently iterative process, relying on the continuity of previous queries, visualization states, and complex reasoning steps. However, to maximize horizontal scale and multi-tenant isolation, AI providers architect reasoning engines as stateless functions [8]. This places the burden of ``memory'' entirely on the application tier, requiring a \textit{Full-History Round-Trip} model where the complete conversation history and metadata must be re-transmitted with every query.

In practice, this model creates three primary areas of operational friction:
\begin{itemize}
    \item \textbf{The Governance Gap:} When organizations rely on centrally-managed session models (where the AI vendor stores history), they lose sovereign control over sensitive conversational logs. This restricts the organization's ability to enforce granular, platform-level security policies or adhere to regional data residency laws.
    \item \textbf{The Caching Conflict:} Modern inference infrastructure utilizes KV (prefix) caching to reduce latency. Naive ``hydration'' of requests often includes rotating security tokens or dynamic metadata at the start of a prompt, which effectively ``busts'' the provider-side cache and increases token costs.
    \item \textbf{Semantic Fragmentation:} Precise analytical reasoning requires high-fidelity grounding in data schemas. Passing raw metadata from a client-side wrapper is both insecure and inefficient, particularly for large-scale enterprise schemas with thousands of fields.
\end{itemize}

% --- TABLE 1 ---
\begin{table*}[t]
\footnotesize
\caption{Comparison of Architectural Archetypes}
\label{tab:comparison}
\begin{tabularx}{\textwidth}{l X c X}
\toprule
\textbf{Archetype} & \textbf{State Ownership} & \textbf{Model Agnostic} & \textbf{Grounding Strategy} \\
\midrule
Naive Wrapper & Shared between client and shared server state. & Yes & Static schema injection; prone to exhaustion. \\
Centrally-Managed & Offloaded to the AI vendor's proprietary cloud. & No & Naive RAG managed by provider infrastructure. \\
Ecosystem-Coupled & Tethered to a specific data platform or ecosystem. & Limited & Tightly-bound to vendor-specific metadata. \\
Self-Hosted & Resident on private organizational infrastructure. & Yes & Requires external state and grounding management. \\
\textbf{Hydration Proxy Pattern} & \textbf{Sovereign ownership within organization storage.} & \textbf{Yes} & \textbf{Cache-Aware Multi-Stage Expansion.} \\
\bottomrule
\end{tabularx}
\end{table*}

\section{The Architectural Framework: The Hydration Proxy}
To resolve these frictions, we present a three-tier architecture synthesized from the requirements of production-grade analytical systems. This framework establishes a clear separation between the UI, the security gateway, and the persistence layer.

\begin{itemize}
    \item \textbf{Orchestration Layer (Frontend):} Serves as the primary entry point, maintaining the immediate UI state and reassembling the chronological message history required for each request.
    \item \textbf{Hydration Proxy (Secure Gateway):} Acts as a specialized backend bridge that ``hydrates'' raw client requests with authoritative platform context. This tier is responsible for the dynamic injection of security credentials and governing metadata (e.g., schemas) that cannot reside on the client tier.
    \item \textbf{Hybrid Persistence Layer:} Employs a dual-storage strategy that decouples session management from conversational content, ensuring the system can scale to handle dense, high-volume interaction logs.
\end{itemize}

\section{State Management: Decoupled Persistence}
Conversational history in analytical systems is information-dense, containing large JSON payloads with query results and visualization metadata. We identify a decoupled strategy that ensures high-throughput access without compromising database performance:

\begin{itemize}
    \item \textbf{Relational Ledger (Metadata Tier):} A relational database maintains the structured ``skeleton'' of the session. This ledger tracks session ownership, the chronological sequence of messages, and unique pointers to the full conversation data. This allows the system to perform fast lookups and audit session activity without loading massive text blobs.
    \item \textbf{Object Storage Tier (Content Tier):} High-volume, unstructured payloads are offloaded to an encrypted object storage service. This prevents relational database bloat and ensures that the system can scale as users generate thousands of data-dense conversational turns.
    \item \textbf{Full-History Reassembly:} Upon every user interaction, the proxy uses the relational ledger to locate the relevant content in object storage. It reassembles the sanitized history, ensuring the reasoning engine has the persistent context required to resolve follow-up questions.
\end{itemize}

\section{Comparative Systems Analysis}
We evaluate the architectural trade-offs of this pattern against prevailing industry archetypes:

\textbf{Naive Stateless Wrappers:} Common in initial prototypes, this involves utilizing standard stateless endpoints, such as the OpenAI Chat Completions API [7]. While application-tier memory libraries attempt to manage context recall via client-side summarization loops, they are prone to severe context bloat and introduce security vulnerabilities by exposing raw platform metadata and governance boundaries directly to the client tier [4].

\textbf{Centrally-Managed Session Models:} Providers like the OpenAI Assistants API [1] store history and manage context windowing on proprietary infrastructure. While this minimizes implementation complexity, it presents challenges for enterprise data sovereignty and multi-cloud portability.

\textbf{Ecosystem-Integrated Agent Models:} Analytical platforms like Databricks Genie [2], Snowflake Cortex [3], or Gemini in BigQuery [9] utilize a tightly-coupled integration model. In these systems, the AI reasoning engine is an intrinsic component of the data warehouse ecosystem. While this enables performance, it creates significant vendor lock-in and restricts multi-cloud flexibility.

\section{The Context Stabilization Mandate}
A substantive design challenge of the Hydration Proxy is the tension with provider-side KV caching. Modern inference infrastructure [6] utilizes KV caching to store attention states for static prefixes. Naive hydration---injecting rotating security tokens or dynamic metadata at the prompt origin---busts this cache, increasing latency and cost.

We identify the \textit{Linear Context Strategy} as a core requirement for production proxies. The Proxy must maintain a stable prompt structure, ensuring that repeated context remains identical for every request:
\begin{enumerate}
    \item \textbf{Static Prefix:} High-token platform schemas and system instructions are fixed at the prompt origin to maximize cache reuse across multiple turns.
    \item \textbf{Semi-Stable History:} Chronological logs are reassembled without mutation to preserve byte-identity across turns, ensuring the cache is only appended, never invalidated.
    \item \textbf{Volatile Suffix:} Ephemeral data, including rotating OAuth tokens, user location metadata, and the current query, are relegated to the suffix.
\end{enumerate}

\section{System Merits and Design Synthesis}
The architectural merits of the \textit{Hydration Proxy Pattern} are synthesized below:

\begin{itemize}
    \item \textbf{Sovereign Persistence:} By offloading message payloads to an organization-controlled storage tier, conversational state remains platform-resident. This ensures users can resume complex reasoning context across multiple devices without involving the AI vendor's storage.
    \item \textbf{Governed Semantic Grounding:} The proxy tier prevents sensitive platform schemas from being exposed to the client. This allows for the dynamic injection of authoritative field-level security constraints just-in-time for the reasoning task, ensuring the model is grounded only in data the user is authorized to access.
    \item \textbf{Responsive Interleaving:} Through real-time stream parsing at the proxy tier, the system can separate the model's internal reasoning steps (thoughts) [5] from user-facing answers. This enables the UI to drive immediate progress states, mitigating the inherent latency of complex analytical tasks.
 \item \textbf{Strategic Context Scaling:} The Proxy acts as a ``context governor,'' implementing \textit{Recursive Summarization} and
     \textit{Removable Reasoning}. By periodically compacting history or stripping high-token internal reasoning steps [5], the Proxy
     maintains operational reliability. While compaction causes a transient cache miss, it resets the prompt's token baseline, ensuring the
     session remains viable as conversational density increases.
\end{itemize}

\section{Conclusion}
By identifying the \textit{Hydration Proxy Pattern} and the \textit{Context Stabilization Mandate}, we can architect Conversational Data Systems that are scalable, secure, and provider-agnostic. This architecture allows platforms to maintain sovereign control over the ``Memory'' and ``Governance'' of the system while utilizing stateless reasoning APIs for ``Intelligence.'' It resolves the tension between data privacy and the infrastructure-level economic necessity of prefix caching.

\end{document}